\documentclass[letterpaper, 10 pt, conference]{ieeeconf} 

\usepackage{tabularx,booktabs}
\usepackage[flushleft]{threeparttable}
\usepackage{nameref}
\usepackage{amsmath}
\usepackage[utf8]{inputenc}
\usepackage{booktabs}
\usepackage{cite}
\usepackage{graphicx}
\usepackage{color}
\usepackage{url}
\usepackage{cite}
\usepackage{comment}
\usepackage{amsfonts}
\usepackage[per-mode=symbol]{siunitx}
\usepackage{mathtools}
\usepackage{setspace}
\usepackage{bm}
\usepackage{pifont}

\IEEEoverridecommandlockouts          
\title{\LARGE \bf StaccaToe: A Single-Leg Robot that Mimics the Human Leg and Toe}

\author{Nisal Perera$^{1}$, Shangqun Yu$^{1}$, Daniel Marew$^{1}$, Mack Tang$^{2}$, Ken Suzuki$^{3}$, Aidan McCormack$^{3}$, \\Shifan Zhu$^{1}$, Yong-Jae Kim$^{4}$, and Donghyun Kim$^{1}$
\thanks{$^{1}$ College of Information and Computer Sciences, University of Massachusetts Amherst, MA, U.S. ({\tt\small donghyunkim@cs.umass.edu}) }
\thanks{$^{2}$ University of Maryland College Park}
\thanks{$^{3}$ College of Engineering, University of Massachusetts Amherst, MA, U.S.}
\thanks{$^{4}$ Korea University of Technology and Education (KOREATECH), Cheonan-City, Rep. of Korea ({\tt\small yongjae@koreatech.ac.kr})}%
}

\begin{document}

\maketitle
\thispagestyle{empty}
\pagestyle{empty}

\begin{abstract}
We introduce StaccaToe, a human-scale, electric motor-powered single-leg robot designed to rival the agility of human locomotion through two distinctive attributes: an actuated toe and a co-actuation configuration inspired by the human leg. Leveraging the foundational design of HyperLeg's lower leg mechanism, we develop a stand-alone robot by incorporating new link designs, custom-designed power electronics, and a refined control system. Unlike previous jumping robots that rely on either special mechanisms (e.g., springs and clutches) or hydraulic/pneumatic actuators, StaccaToe employs electric motors without energy storage mechanisms. This choice underscores our ultimate goal of developing a practical, high-performance humanoid robot capable of human-like, stable walking as well as explosive dynamic movements. In this paper, we aim to empirically evaluate the balance capability and the exertion of explosive ground reaction forces of our toe and co-actuation mechanisms. Throughout extensive hardware and controller development, StaccaToe showcases its control fidelity by demonstrating a balanced tip-toe stance and dynamic jump. This study is significant for three key reasons: 1) StaccaToe represents the first human-scale, electric motor-driven single-leg robot to execute dynamic maneuvers without relying on specialized mechanisms; 2) our research provides empirical evidence of the benefits of replicating critical human leg attributes in robotic design; and 3) we explain the design process for creating agile legged robots, the details that have been scantily covered in academic literature.
\end{abstract}

\section{Introduction}
Our overarching ambition is to develop a humanoid robot that mirrors human locomotion agility. Agile movements like jumping offer a flexible and dynamic mode to navigate over challenging terrains, which allows us to overcome barriers, evade danger, and traverse disconnected grounds. A robot with such agility will be able to expand its operational range across different terrains and reach elevated vantage points for surveillance and monitoring. To establish an important milestone toward human-level mobility in robots, we introduce StaccaToe, a single-leg robot designed to perform both stable balance control and explosive jumping motion. 

\begin{figure}
    \centering
    \includegraphics[width=\linewidth]{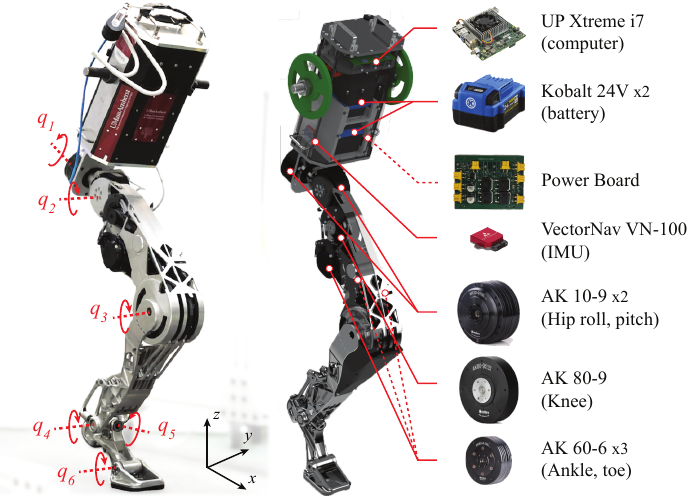}
    \caption{{\bf StaccaToe robot and component explanation.} 
StaccaToe features six actuators, including a two-DoF ankle and a toe. The robot is designed as a standalone system, having an onboard control computer, sensors, and power system.}
    \label{fig:overview}
    \vspace{-0.5cm}
\end{figure}

In jumping robot development, hydraulic and pneumatic actuators have been widely adopted because of their high force/torque density, rapid response time, and impact robustness~\cite{hyon2002development, HSLR, mowgli, HyQ, pneuSpider, softPneu}. An early work of Marc Raibert et al.~\cite{marc3Dhop} investigated a 3D hopping robot that consisted of a hydraulically actuated hip and a leg using a pneumatic actuator. Recently, Atlas from Boston Dynamics and Festo Bionic-Kangaroo~\cite{festoBionicKang} achieved jumping through hydraulic and pneumatic actuation, respectively. However, systems using hydraulic actuation suffer from inefficiency and maintenance issues such as oil leaks~\cite{hydExo, surveyHy}. Pneumatic systems are also inefficient, alongside other issues such as limited energy storage, low precision, and noisy operation.

On the other hand, significant efforts have also been made to build agile robots using electric actuators. However, most robots have utilized specially designed mechanisms to overcome the limited output torque of electric motors. For instance, Salto-1P~\cite{salto} achieved jumps exceeding $1~\si{\meter}$ in height by integrating a series elastic actuator along with a variable mechanical-advantage limb. Other small-scale jumping robots such as TAUB~\cite{TAUB}, JumpRoACH~\cite{jumpROACH}, GRILLO III~\cite{grillo3}, and ARCHER~\cite{ambrose2022creating} also leverage mechanical advantages to enhance jumping performance. There are also some human-scale jumping robots like RAMIEL using a parallel wire-driven mechanism~\cite{suzuki2022ramiel}. While these accomplishments are impressive, it is nontrivial to extend the mechanisms specially designed for jumping to general-purpose legged robots. Moreover, specialized mechanisms such as serial springs~\cite{ambrose2022creating, grimes2012design} or tendon-driven winding mechanisms~\cite{TAUB} can hinder stable nominal locomotion, although they are beneficial for certain types of motion.


There exist other robots that are not designed for hopping but are capable of showing impressive jumping. However, many of them are quadruped robots that can utilize multiple, relatively short legs~\cite{cheeta3jump,spaceBOKjump,minitaur} or assisted by a constraining mechanism~\cite{KURMET, hoppy}, or their jumping height is not comparable to that of a human's~\cite{BRL, Ascento, cassieRL}. Recent developments have seen the commercial release of humanoid robots equipped with electric motors that possess the ability to jump \cite{Unitree}. However, detailed methodologies behind these capabilities remain undisclosed to the public.  In conclusion, prior studies have not yet presented a solution to accomplish both controlled nominal locomotion and explosive dynamic movements in a human-scale, electric motor-powered biped robot. 

\begin{figure}[t!]
    \centering
    \includegraphics[width=\linewidth]{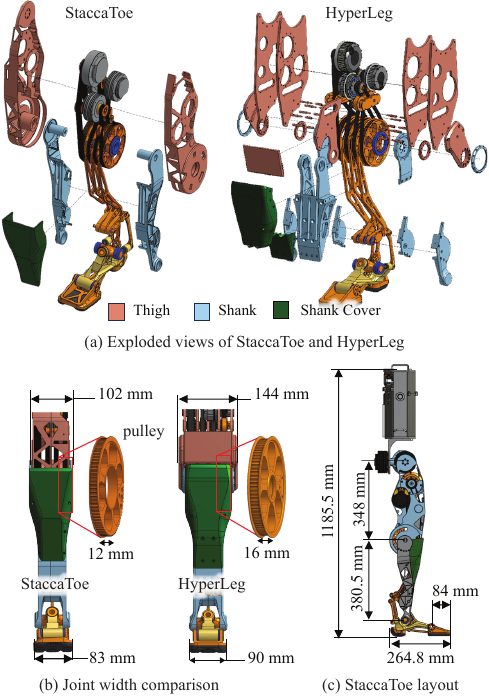}
    \caption{ {\bf Design comparisons between StaccaToe and HyperLeg, and the overall dimensions of StaccaToe.} (a) The number of components in StaccaToe is significantly fewer than that in HyperLeg. This results in reduced maintenance, a lighter weight, and enhanced stiffness. (b) The width of StaccaToe's leg is reduced by 30\% by simplifying the pulley design. (c) StaccaToe is designed to be similar to the average human leg proportions~\cite{NASAanthro}.}\label{fig:hardware}
\end{figure}

We propose a new hopping robot, named StaccaToe: a human-scale single-leg robot that can perform stable balance control as well as explosive jump. It stands approximately $1.2~\si{\meter}$ tall (Fig.~\ref{fig:hardware}(c)), weighs $16~\si{\kilogram}$ and equips 6 electric motors. StaccaToe has two unique features inherited from HyperLeg~\cite{hyperleg}: co-actuation and actuated toe mechanism. 
To enable high-force exertion, we carefully designed the drivetrains of knee, ankle, and toe actuators to assist the knee during leg extension. The cooperative actuation~\cite{tello} scheme enables StaccaToe to generate large ground reaction forces that cannot be accomplished if the actuators are configured serially or coupled in a way that the actuators bother each other's motion during push-off. We formulate trajectory optimization to fully exploit the co-actuation setup and accomplish the jumping of StaccaToe experiments thanks to the augmented knee torque output. 

Another unique feature of StaccaToe is an actuated toe. Despite compelling evidence that underscores the toe's pivotal role in both human and robotic movement~\cite{chou2009role, toeathletic, wang2006influence}, toe mechanisms in robots are often overlooked, primarily due to their mechanical complexity and vulnerability to impacts. Although several toe mechanisms have been proposed, they are either passive, lacking the ability to provide propulsive force or active balancing~\cite{ogura2006human, hashimoto2010study}, or actuated but heavy and prone to damage, rendering them unsuitable for dynamic locomotion~\cite{sato2010trajectory, choi2016design, yamamoto2007toe}.
Unlike the prior designs, StaccaToe's toe is light and robust to external impacts while maintaining high control fidelity to enable balance control of a floating base. Our tiptoe balance experiment results demonstrate that the overall drivetrains including a toe offer sufficiently stiff control to maintain its balance. 

In summary, this paper's main contributions encompass the following key aspects: 1) development of a human-scale, stand-alone floating-base hopping robot, StaccaToe, through new link design and extensive design optimization, 2) experimental validation of the effectiveness of actuated toe and co-actuation mechanisms by demonstrating tiptoe balance and jumping, and 3) documentation of detailed design processes and challenges associated in dynamic legged robot design and control (e.g., actuator identification, cable management, power electronics).





\section{Robot Hardware Design and Development}
\label{sec:robot}

The HyperLeg~\cite{hyperleg} showcased a leg design inspired by human biomechanics, integrating important features for acrobatic maneuvers such as an actuated toe, cooperative actuation~\cite{tello}, non-collocated actuators, and a large range of motion. While promising, the robot consists only of the lower leg part, lacking hip joints and a body, and its experiments required the assistance of a planar constraint mechanism. As HyperLeg serves as a test platform to demonstrate the feasibility of two features -- the actuated toe and co-actuation mechanisms -- several details like linkage stiffness, power electronics, body design, mass reduction, and balance control were not included in its development. In this paper, we create a standalone robot by 1) refining the original design to reduce component count, 2) optimizing the durability of linkages while minimizing weight, 3) conducting extensive analyses of actuator parameters, 4) developing custom power electronics, and 5) configuring mechanically safe cable/connector management.

\subsection{Reduction in component count and leg width}
StaccaToe comprises seven primary modules: torso, hip, thigh, shank, ankle, foot, and toe. Among those, the thigh and shank modules are especially important because they need to endure primary loading during locomotion and encapsulate various components such as transmission links, pulleys, and actuators. As depicted in Fig.~\ref{fig:hardware}(a), the thigh and shank links of HyperLeg consist of multiple parts, necessitating many fasteners such as screws and washers. To be a sustainable system, it is essential to trim the number of components, thereby reducing potential failure points and maintenance difficulties. In the design of StaccaToe, we amalgamated several components into two links for both the thigh and shank. This consolidation has reduced the component count for thigh and shank by 28 and 7, respectively. 

Fig.~\ref{fig:hardware}(b) shows the widths of the knee and ankle joints in both models. In our redesigned leg, the widths of the knee and ankle are trimmed down by $42~\si{\milli\meter}$ and $7~\si{\milli\meter}$, respectively. This not only results in a significant weight reduction but also a streamlined leg form factor. Such enhancements will be beneficial for preventing self-collision when we extend this leg to bipedal robots.

\subsection{Topology Optimization of Primary Links}
\begin{figure}
    \centering
    \includegraphics[width=\linewidth]{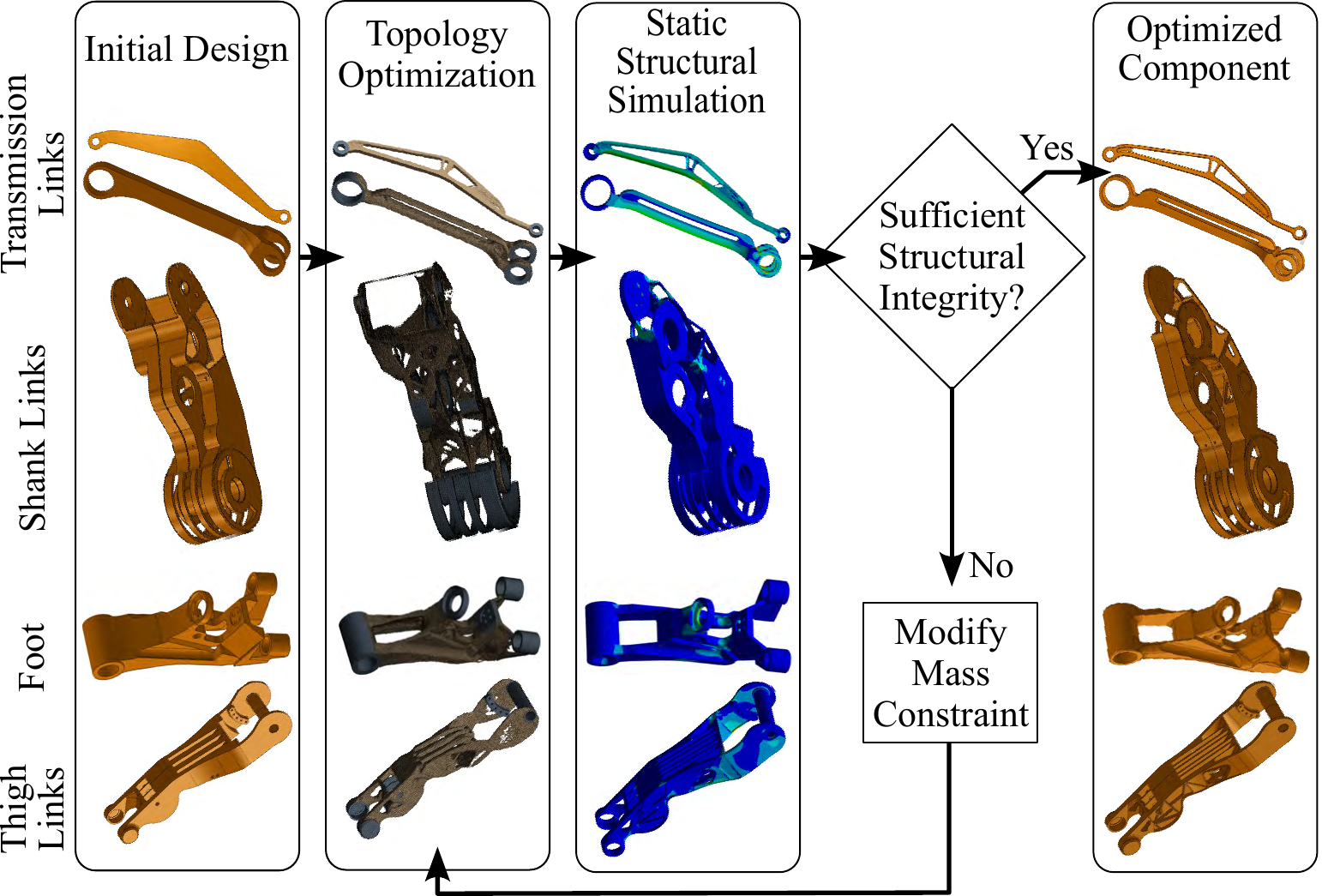}
    \caption{{\bf The topology optimization process.} Optimized components are reduced in mass while maintaining structural rigidity.  }\label{fig:topology_opt}
\end{figure}

To reduce link mass without compromising structural rigidity, we employed topology optimization on important components:  thigh, shank, foot, and power transmission links. We used ANSYS~\cite{ANSYS} for topology optimization to refine material distribution within a predefined design space. Fig.~\ref{fig:topology_opt} illustrates this optimization process, aiming to minimize strain energy while satisfying the mass reduction constraint. Minimizing strain energy in optimization ensures structural integrity and leads to optimized structures with better performance characteristics such as increased stiffness, reduced deflection, and improved natural frequencies. During the process, we employed Sequential Convex Programming as the primary solver and subsequentially simulated static structure to assess the structural integrity of components with the refined topology. These optimized components are designed to withstand impact forces up to $600~\si{\newton}$, or roughly four times the total weight of StaccaToe. Additionally, the shank and thigh links can endure twisting moments up to $50~\si{\newton\meter}$ along their primary axes.

Throughout the design optimization, StaccaToe's leg achieved a mass reduction of approximately $0.47~\si{\kilogram}$ compared to HyperLeg while maintaining structural rigidity. The total mass of the lower leg components (shank, ankle, and foot) was reduced by 14.78\%, dropping from $3.79~\si{\kilogram}$ to $3.23~\si{\kilogram}$. Although StaccaToe's thigh link has a slight increase in weight, measuring $4.03~\si{\kilogram}$ against HyperLeg's $3.94~\si{\kilogram}$, we considerably decreased the component count with a net reduction of 28 components. As depicted in Fig.~\ref{fig:hardware}(a), HyperLeg's thigh comprises four aluminum plates connected by metal axles, offering lightness but at the cost of reduced stiffness. In contrast, StaccaToe's thigh, with an increment of $0.09~\si{\kilogram}$ and enhanced design, has the capability to withstand a torsional torque of $50~\si{\newton\meter}$ and a compression force four times its own weight.


\subsection{Motor Control and Power System}
Staccatoe's actuators are powered by MAB Robotics MD80 V2.1 motor controllers, which replaced the T-Motor's original driver boards. Communication between the UP Xtreme Intel Core i7 controller and the motor controllers is facilitated by two MAB Robotics CANdle devices, with each CANdle device serially connecting to three actuators. The CANdle devices utilize USB 3.0 for communication with the controller. We were able to implement a real-time communication system that runs at $500~\si{\hertz}$ utilizing CAN FD. 

\begin{figure}[t]
    \centering
    \includegraphics[width=\linewidth]{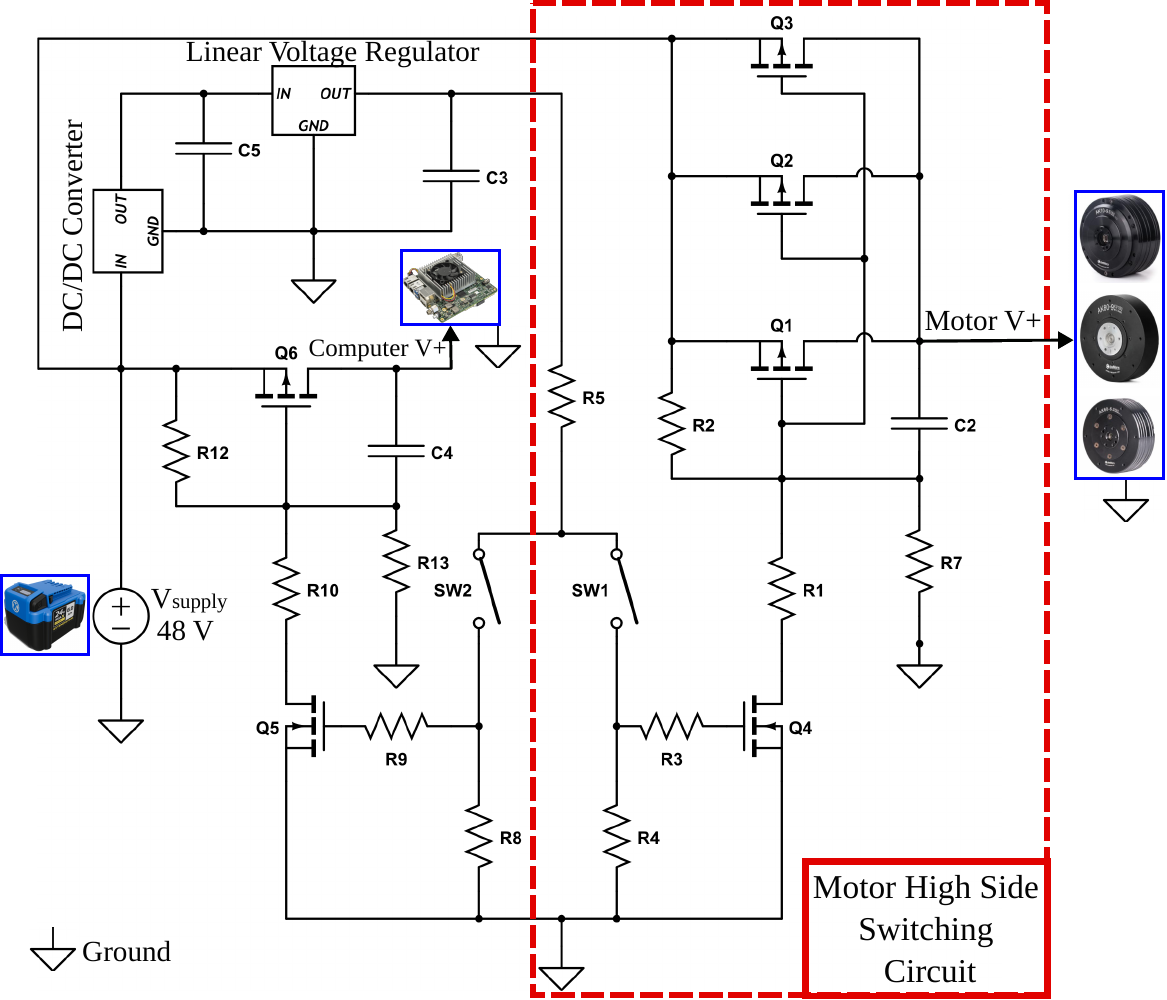}
    \caption{{\bf Switching circuitry.} The primary high side switching circuitry is created to provide power to actuators and controller.  }\label{fig:pwbd}
\end{figure}

For the power system, we engineered a custom powerboard capable of delivering high currents at voltages up to $48~\si{\volt}$. As will be evident from our trajectory optimization results in the later section, StaccaToe has the potential to perform agile maneuvers, such as jumping, by fully exploiting the capabilities of its actuators. To reach the peak torques of the actuators, a power system capable of providing high currents is required. To achieve this, we implemented a high side switch using three parallel IXTT140P10T P-channel MOSFETs, each rated for a continuous drain current of $140~\si{A}$, as highlighted in Fig.~\ref{fig:pwbd}. Additionally, we utilized a similar high side switch configuration to provide power to the controller. Furthermore, operating at $48~\si{\volt}$ (the maximum allowable voltage of the MD80 motor controllers) allows us to increase the speed limits of the actuators. The operating voltage was achieved by connecting two $24~\si{\volt}$ Kobalt Li-ion batteries in series to create a $48~\si{\volt}$ power source. 

\subsection{Actuator Identification}
\begin{figure}[t]
    \centering
    \includegraphics[width=\linewidth]{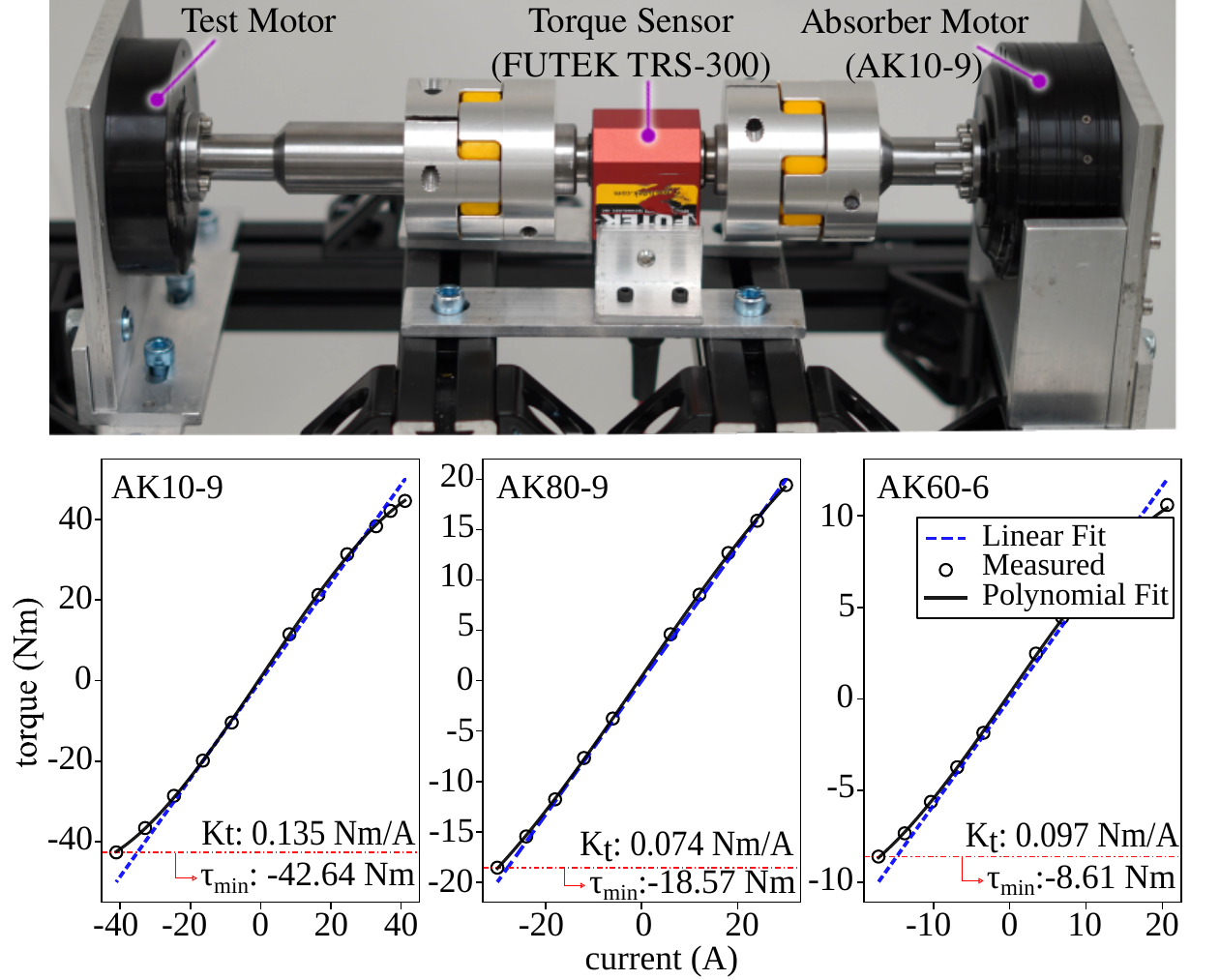}
    \caption{{\bf Motor torque characteristics.} The relationship between the torque and current of each actuator is measured using our dynamometer (top). The measured torque constant and peak torque for each actuator are as specified in the plot.} 
    \label{fig:dyno}
\end{figure}

We identified the actual torque constants, $K_t$, and peak torques, by measuring these parameters of the AK10-9, AK80-9, and AK60-6 actuators using our dynamometer (see Fig. \ref{fig:dyno}). It is worth noting that the measured torque constant values were lower than those specified in the actuator specifications. These measurements are utilized in the actuator-level impedance control to ensure proper commanded torque generation.
Another crucial aspect of the actuator is its backlash. Given StaccaToe's scale, significant backlash in the knee actuator can lead to substantial errors in kinematic computations. To measure the backlash of the AK80-9 knee actuator, we rigidly fixed the output shaft and recorded the motor encoder data while commanding torques. Our experiments revealed a 0.15-degree backlash at the actuator output, which was within the manufacturer's backlash specification of 0.19 degrees. Given the 40/9 reduction ratio between the knee actuator output and the knee joint, and a knee-to-body length of $435~\si{\milli\meter}$, 0.19 degrees of backlash in the knee actuator corresponds to about $6.4~\si{\milli\meter}$ of error in the robot body position. It was observed in a previous investigation on a similarly scaled robot~\cite{passiveankle} that a deviation of about $10~\si{\milli\meter}$ in body position due to backlash still yields acceptable results.

\begin{figure}[t!]
    \centering
    \includegraphics[width=\linewidth]{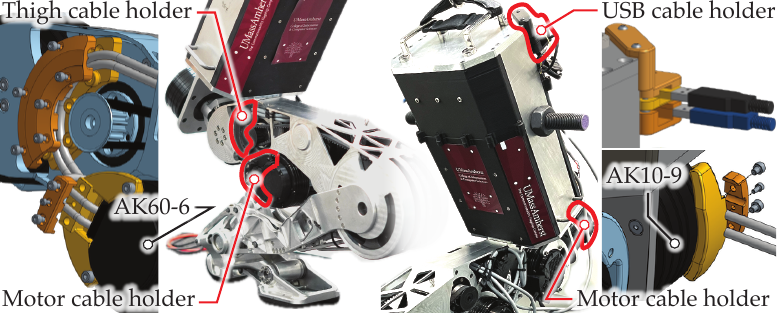}
    \caption{{\bf Cable holders} located at vulnerable points of the robot, ensuring reliable power and signal connections.} \label{fig:cable_mgmt}
\end{figure}

\subsection{Cable Management}
Wiring and cable management is often regarded as one of the most troublesome aspects of electro-mechanical systems. Strategic wiring is crucial to minimize the risk of wires getting entangled with their surroundings or tampering with the mechanical assembly. It is also important to rigidly hold wire connections, because they are susceptible to becoming loose, which causes electrical connectivity issues. Problems in electrical connectivity can cause signal noise, communication loss, and malfunction in feedback control, which may eventually lead to catastrophic hardware failure. To avoid these potential pitfalls, custom cable holders were designed to tightly hold the signal and power cables. As shown in Fig.~\ref{fig:cable_mgmt}, the custom cable holders were fixed near the vulnerable motor and computer connections. Screw-on covers on the cable holders secure the cables by firmly compressing them into a carefully tailored cylindrical groove. This ensures minimal cable slack and stable connections to the motors.

\section{Jumping Trajectory Optimization}

In this study, we employ trajectory optimization based on a single rigid body dynamics model coupled with full kinematics to generate jumping trajectories~\cite{chignoli2021humanoid, dai}. One key motivation for adopting this streamlined approach in this work instead of employing full-body dynamics trajectory optimization or centroidal dynamics is that this approach considerably simplifies the complexity of nonlinearities in the optimization problem. This strategic simplification primarily aims to accommodate the intricate nonlinearities introduced by the co-actuation mechanism. \cite{chignoli2021humanoid, dai}.

\begin{figure}
    \centering
    \includegraphics[width=\linewidth]{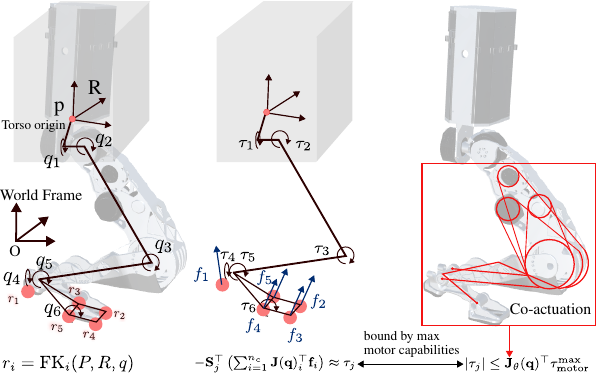}
    \caption{{\bf Optimization model.} (Left) We approximate the center of mass of the single rigid body $\mathbf{p}$ and orientation $\mathbf{R}$ to be aligned with the robot's torso's frame. Forward kinematics functions $\mathrm{FK}_i$ are derived to map torso's origin, orientation and joint positions to contact position $r_i$. During the contact phase, the contact point positions are constrained to be fixed on the ground to ensure kinematic feasibility. (Middle) The contact jacobains are derived to map reaction forces to joint torque. (Right) The co-actuation Jacobian $J_\theta$ is the mapping from motor velocities to joint velocities. By utilizing the co-actuation jacobian, we could enforce motor-level torque constraints instead of conservative, approximated ground reaction force-level or joint torque-level limits.} \label{fig:optimization}
\end{figure}

In our trajectory optimization formulation, the single rigid body dynamics of the robot relate the ground reaction forces to the robot's linear and angular momentum, while the kinematics model is used to determine a corresponding kinematic trajectory consistent with the center of mass (CoM) and contact locations. The optimization encompasses the following state variables:

  \begin{equation}
     \mathbf{x} = [\mathbf{p^\top} \quad \mathbf{R_{vec}} \quad \mathbf{v^\top} \quad \bm{\omega^\top} ]^{\top},
 \end{equation}
where $\mathbf{p} \in \mathbb{R}^3$ and $\mathbf{v} \in \mathbb{R}^3$ are the position and the velocity of the robot's center of mass. $\mathbf{R_{vec}} \in \mathbb{R}^9$  and $\bm{\omega} \in \mathbb{R}^3$ are the vectorized orientation matrix and angular velocity of the body frame. The optimization is formulated by
\begin{equation*}
\min_{\mathbf{x}_k} \quad \sum_{k=1}^{N} \ (\mathbf{X}_{err, k}^\top \ \bm{Q} \ \mathbf{X}_{err, k})
\end{equation*}
where $\mathbf{X}_{err, k} = [\mathbf{p}_{err, k}, \mathbf{R}_{err, k}, \mathbf{v}_{err, k} ,\bm{\omega}_{err, k}]$. Except for the orientation error, all other errors are calculated by subtraction between the current and desired states. For the orientation error,  we first define $\bm{R}_{err,k}$ given by
\begin{equation}
     \bm{R}_{des,k} \bm{R}_{err,k} = \bm{R}_k, 
\end{equation} 
where $\bm{R}^{des}_k$ is the desired orientation matrix for the $k$-th step and $\bm{R}_k $ is the current orientation matrix at $k$-th step, thus
\begin{equation}
\bm{R}_{err,k} = {\bm{R}_{des,k}}^\top \bm{R}_k.
 \end{equation}
The skew-symmetric matrix form of the axis for the orientation matrix can be expressed as 
\begin{equation}
[ \hat{\bm{\omega}} ] = 
     \begin{bmatrix}
      0 & -\hat{\omega}_3 & 
 \hat{\omega}_2 \\ 
      \hat{\omega}_3 & 0 & -\hat{\omega}_1 \\
      -\hat{\omega}_2 & \hat{\omega}_1 & 0
    \end{bmatrix} 
    = \frac{1}{2 \sin{\theta}} (\bm{R} - \bm{R^\top}).
 \end{equation}
 Thus 
 \begin{equation}
     \bm{\hat{\omega}}_{err,k} \theta_{err,k} = \left(\frac{\theta_{err,k}}{2 \sin{\theta_{err,k}}} (\bm{R}_{err,k} - \bm{R}^{\top}_{err,k})\right)^{\vee}
 \end{equation}
where $ \mathbf{\hat{\omega}_{err,k} } \theta^{err}_k \in \mathbb{R}^3$ is the orientation error, $( \cdot)^\vee : \mathfrak{so}(3) \rightarrow \mathbb{R}^3$ is the inverse of the skew function. To enhance the efficiency of the trajectory optimization, we approximate $\theta_{err,k}$ by $\sin{\theta_{err,k}}$ when $\theta_{err,k}$ is small. Our orientation error is given as 
  \begin{equation}
     \bm{\hat{\omega}}_{err,k} \theta_{err,k} = \left(\frac{1}{2} (\bm{R}_{err,k} - {\bm{R}_{err,k}}^\top)\right)^{\vee}.
 \end{equation}
The constraints are given as follows: 
\begin{align*}
  & \mathbf{x}_{k+1} = \mathrm{dynamics}(\mathbf{x}_k, \triangle t) &  \text{\ding{172}} \\
  &  \mathrm{FK}_i(\mathbf{p}_k, \mathbf{R}_k, \mathbf{q}_k) = \mathbf{r}_{i,k}  &  \text{\ding{173}} \\
  &  \mathbf{r}_{i,k} c_{i,k} = \mathbf{r}_{i,k+1}c_{i,k}   &  \text{\ding{174}} \\
  & -\mu f_{i,k}^{x || y} \leq f_{i,k}^{z} \leq \mu f_{i,k}^{x || y},  &  \text{\ding{175}}  \\
  & f_{i,k}^z (1 - c_i) = 0, &  \text{\ding{176}}\\
  & \mathbf{q}_{\mathrm{min}} \leq \mathbf{q}_k \leq \mathbf{q}_{\mathrm{max}}, &  \text{\ding{177}}\\
  & | \mathbf{\tau}_{j,k} | \leq  \mathbf{J}_{\theta}(\mathbf{q}_k)^\top \mathbf{\tau}_{\mathrm{motor}}^{\mathrm{max}}   &  \text{\ding{178}}
\end{align*}
where \ding{172} is the dynamics constraint. \ding{173} is the forward kinematics constraint for the contact points, where $\mathbf{q}_k$ is the joint position and $\mathbf{r}_{i,k}$ is the $i$-th contact point's position. \ding{174} enforces the non-slipping constraint, where $c_{i,k}$ indicates if the $i$-th contact point is under contact or not. \ding{175} is the friction cone constraint. \ding{176} is the contact force constraint. \ding{177} is the joint limit constraint. Finally, \ding{178} is the motor torque constraint to ensure that the generated trajectories comply with the robot's actuator limits. As noted in previous work \cite{chignoli2021humanoid}, which our observations also support, the torques needed to realize these trajectories are largely dominated by those required to generate ground reaction forces. Therefore, in this work, we estimate the necessary torque as equivalent to that required for producing ground reaction forces and establish limits accordingly, which is given by
\begin{equation}
\mathbf{\tau}_{j,k} \approx - \mathbf{S}_j^\top \left( \sum_{i=1}^{n_c} \mathbf{J}(\mathbf{q}_k)_{i}^\top \mathbf{f}_{i,k} \right),
\end{equation}
where \(\mathbf{\tau}_{j, k} \in \mathbb{R}^{n}\) is the required joint torque at $k$-th step, \(\mathbf{S}_j \in \mathbb{R}^{n\times n+6}\) is a selection matrix, and \(\mathbf{J}_{i}\in \mathbb{R}^{3\times n+6}\) is the Jacobian of the $i$-th contact point.  Due to the special co-actuation design of StaccaToe, the maximum torque that can be applied at a joint is dependent on the robot's configuration. Consequently, we cannot directly apply torque limit constraints using simple upper and lower bounds as done in prior studies \cite{chignoli2021humanoid}. Instead, we use the following constraints:
\begin{align}
\label{eq:torque_stance_cons}
|\mathbf{\tau}_{j,k} | & \leq \mathbf{J}_{\theta}(\mathbf{q}_k)^\top \mathbf{\tau}_\mathrm{motor}^{\mathrm{max}},
\end{align}
where \(\mathbf{J}_{\theta} \in \mathbb{R}^{n \times n}\) is the motor Jacobian that relates joint velocity to motor velocity.


\section{Results and Discussion}
We used our custom-developed dynamics engine, DARoS-Core, for physics simulations to validate and refine our controllers before deploying them on the robot. We utilize whole-body impulse control (WBIC) proposed in \cite{kim2019highly} to compute the joint position, velocity, and torque commands. For StaccaToe control, we set the default constraints and tasks as highlighted in Table \ref{tb:wbic_setup}.

\begin{table}[t]
\caption{Configuration of WBIC}\vspace{-0.5cm}
\begin{center}
\begin{tabular}{p{3.5cm} p{4.2cm}}
\hline
Contact/Task Name & Description \\
\hline
Point contact constraint & Four points for the toe link/ one for heel \\
Body orientation task & Torso orientation \\
Body position task & Torso position\\
Joint control task & Keep the entire joint posture\\
\hline
\end{tabular}
\label{tb:wbic_setup}
\end{center}
\end{table}


\begin{figure}[t]
    \centering
    \includegraphics[width=\linewidth]{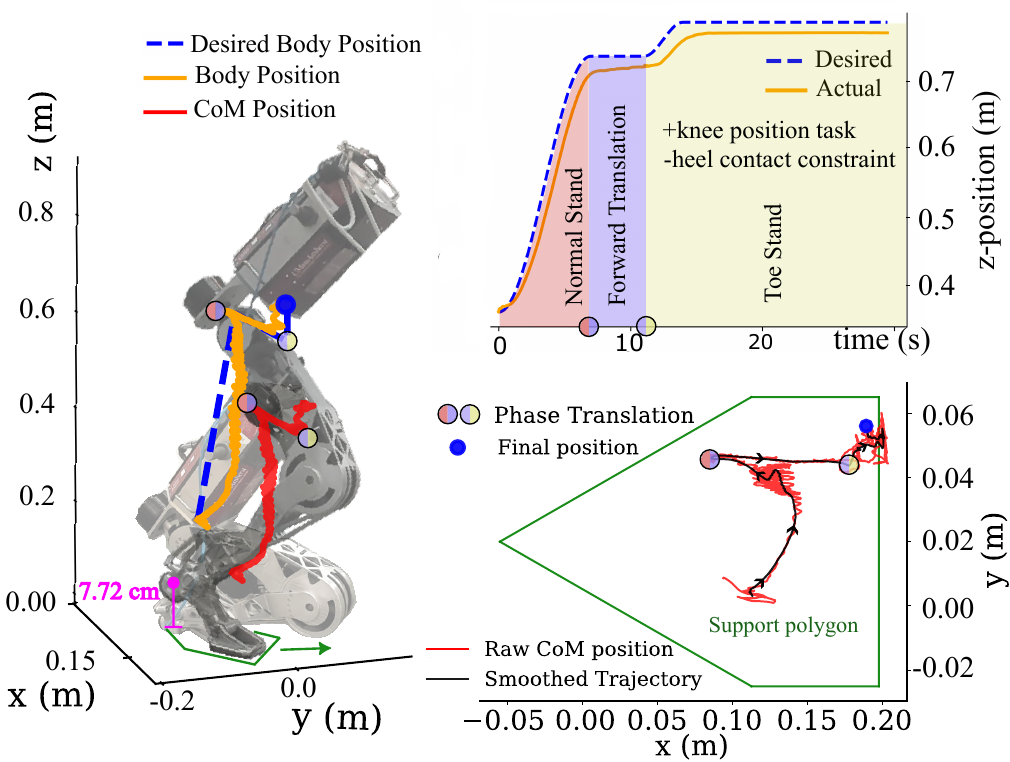}
    \caption{{\bf Tiptoe balance.} WBIC tracking performance during tiptoe stand. The robot's body z-position variation, along with time-based motion phase transition (top right), and CoM position variation in the ground plane (bottom right), are presented.}
    \label{fig:tiptoe}
\end{figure}

\subsection{Tiptoe Balance}

We demonstrate that StaccaToe can perform tiptoe motion using WBIC, maintaining balance by supporting its entire weight on the small footprint of its toe. To achieve tiptoe motion, we introduced a knee joint position task to the WBIC task hierarchy, accompanied by a time-based transition to remove the contact constraint at the heel. As illustrated by Fig.~\ref{fig:tiptoe}, the robot initially rises from a flat-footed stance and moves forward to shift its center of mass onto the toe. Subsequently, we introduce the knee joint position task to stabilize the current joint angle, while removing the contact constraint at the heel. 

For the balance controller, we chose body position control as opposed to CoM position control due to its simplicity. As highlighted in Fig.~\ref{fig:tiptoe}, during the standing-up motion, the robot's CoM remains within the support polygon, maintaining balance, while being robust to the undesired oscillations. These oscillations observed during the motion can be attributed to backlash in the drivetrain, which is a common characteristic in many drivetrain systems that is almost unavoidable. Despite this, our WBIC-based balance controller was able to effectively compensate for these oscillations ensuring dynamic stability. This demonstrates not only the controller's ability to maintain precise coordination and balance but also highlights the drivetrain's sufficient stiffness, which is essential for executing nimble motions.

\begin{figure}[t]
    \centering
    \includegraphics[width=\linewidth]{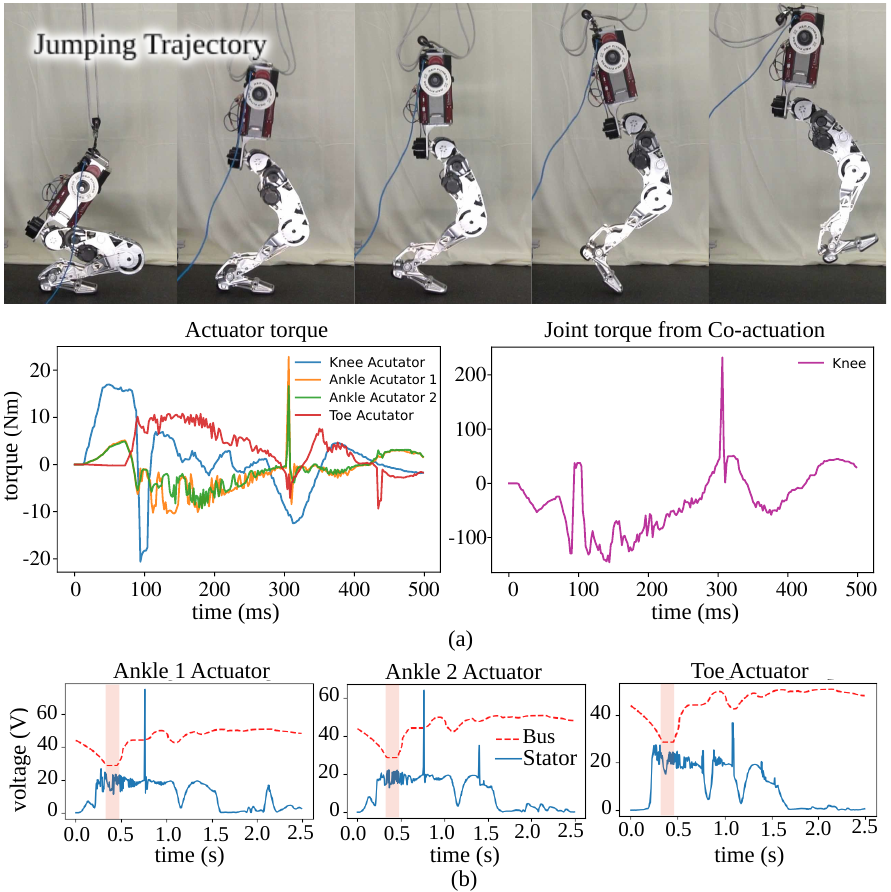}
    \caption{{\bf Vertical jumping.} (a) The figure illustrates the vertical jumping of our StaccaToe robot, achieved through PD control tracking of a pre-defined offline jumping trajectory (top). Through the innovative co-actuation system, the robot's knee joint is able to generate torques significantly beyond the motors' standalone capabilities. (b) The bus and stator voltage profiles during jumping. }\label{fig:jumping_result}
\end{figure}

\subsection{Jumping Test for Evaluation of Co-Actuations}

As previously discussed, co-actuation is a key feature that enhances StaccaToe's agility. This mechanism allows StaccaToe to generate significantly higher torques, particularly at the knee, compared to what individual actuators can produce. To demonstrate this, we performed vertical jumping experiments to assess the effectiveness of co-actuation. We generated a jumping trajectory containing joint angles, velocity and feed-forward torque using the previously outlined trajectory optimization method. This trajectory was then tracked using impedance control with low feedback gains. We added an additional 1~\si{\kilogram} onto the torso of the robot for better momentum transfer. From the results presented in Fig.~\ref{fig:jumping_result}(a),  it is evident that co-actuation has significantly increased torque generation specifically at the knee joint. Specifically, during the push-off phase, the knee joint generated over $100~\si{\newton\meter}$ of torque, exceeding the $80~\si{\newton\meter}$ that the knee actuator alone can produce after the belt reduction. This is approximately about $25\%$ more torque generation. As a result of co-actuation, the knee actuator, ankle actuators, and toe actuator collectively contribute to the overall torque at the knee joint, resulting in an aggregated torque during knee extension. Subsequently, this leads to a higher generation of ground reaction forces. 

Unfortunately, we could not observe the maximum jumping height that we found in our trajectory optimization that reflects actuator torque/velocity limits along with the co-actuation setup. The major constraint comes from the electric power supply because of the batteries' bus voltage drop. As highlighted in Fig.~\ref{fig:jumping_result}(b), during the push-off phase, there is a noticeable drop in bus voltage, accompanied by an increase in stator voltages of the actuators. This effect was particularly prominent in the ankle and toe AK60-6 actuators, since they have to operate at high speeds during push-off. As a result of this, when the difference between the bus and stator voltage diminishes, the torque generation capability of these motors will be reduced momentarily~\cite{chignoli2021humanoid}. Further investigation and hardware updates will mitigate the limitation and fully unlock the physical capability of StaccaToe.




\begin{figure}
    \centering
    \includegraphics[width=\linewidth]{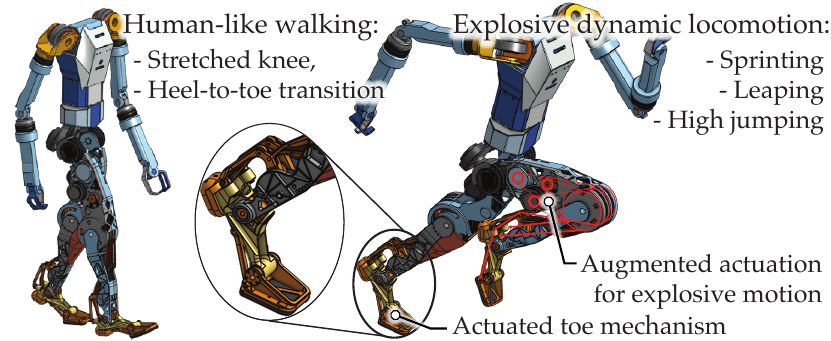}
    \caption{{\bf New humanoid robot, PresToe.} We are building a new humanoid robot by employing the similar co-actuation and toe mechanisms.}
    \label{fig:prestoe}
\end{figure}
\section{Concluding Remarks and Future Plan}

In this paper, we introduce a new human-scale electric motor-driven single-leg robot with an actuated toe and co-actuation. We present the detailed hardware design process including topology optimization, actuator parameter identification, and power electronics. By conducting toe stand and vertical jumping experiments, utilizing WBIC and trajectory optimization respectively, we demonstrate the advantages conferred by the actuated toe and co-actuation. These experiments highlight StaccaToe's remarkable physical capabilities, underscoring its potential for executing agile and explosive movements.

The advancements made in this paper will be incorporated into our new humanoid robot, \emph{PresToe} in Fig.~\ref{fig:prestoe}, the first humanoid robot capable of performing both stable, efficient walking and explosive sprinting and leaping. We have demonstrated that StaccaToe has the physical capabilities to perform intricate motions, such as tiptoe standing, and explosive movements, like jumping, through the use of the actuated toe and co-actuation. Our future plans also involve deploying a hopping locomotion controller to StaccaToe, enabling it to exhibit locomotion capabilities.

\section*{Acknowledgment}
This material is based upon work supported by the National Science Foundation under Grant No. 2220924.

\bibliographystyle{./IEEEtran} 
\bibliography{./IEEEabrv,./reference}

\begin{thebibliography}{10}
\providecommand{\url}[1]{#1}
\csname url@rmstyle\endcsname
\providecommand{\newblock}{\relax}
\providecommand{\bibinfo}[2]{#2}
\providecommand\BIBentrySTDinterwordspacing{\spaceskip=0pt\relax}
\providecommand\BIBentryALTinterwordstretchfactor{4}
\providecommand\BIBentryALTinterwordspacing{\spaceskip=\fontdimen2\font plus
\BIBentryALTinterwordstretchfactor\fontdimen3\font minus \fontdimen4\font\relax}
\providecommand\BIBforeignlanguage[2]{{%
\expandafter\ifx\csname l@#1\endcsname\relax
\typeout{** WARNING: IEEEtran.bst: No hyphenation pattern has been}%
\typeout{** loaded for the language `#1'. Using the pattern for}%
\typeout{** the default language instead.}%
\else
\language=\csname l@#1\endcsname
\fi
#2}}

\bibitem{hyon2002development}
S.-H. Hyon and T.~Mita, ``Development of a biologically inspired hopping robot-``kenken'','' in \emph{Proceedings 2002 IEEE International Conference on Robotics and Automation}, vol.~4.\hskip 1em plus 0.5em minus 0.4em\relax IEEE, 2002, pp. 3984--3991.

\bibitem{HSLR}
X.~Li, H.~Feng, S.~Zhang, H.~Zhou, Y.~Fan, Z.~Wang, and Y.~Fu, ``Vertical jump control of hydraulic single legged robot (hslr),'' in \emph{2019 IEEE/ASME International Conference on Advanced Intelligent Mechatronics (AIM)}, 2019, pp. 1421--1427.

\bibitem{mowgli}
R.~Niiyama, A.~Nagakubo, and Y.~Kuniyoshi, ``Mowgli: A bipedal jumping and landing robot with an artificial musculoskeletal system,'' in \emph{Proceedings 2007 IEEE International Conference on Robotics and Automation}, 2007, pp. 2546--2551.

\bibitem{HyQ}
C.~Semini, N.~G. Tsagarakis, E.~Guglielmino, M.~Focchi, F.~Cannella, and D.~G. Caldwell, ``Design of hyq – a hydraulically and electrically actuated quadruped robot,'' \emph{Proceedings of the Institution of Mechanical Engineers, Part I: Journal of Systems and Control Engineering}, vol. 225, no.~6, pp. 831--849, 2011.

\bibitem{pneuSpider}
A.~Spröwitz, C.~Göttler, A.~Sinha, C.~Caer, M.~U. Öoztekin, K.~Petersen, and M.~Sitti, ``Scalable pneumatic and tendon driven robotic joint inspired by jumping spiders,'' in \emph{2017 IEEE International Conference on Robotics and Automation (ICRA)}, 2017, pp. 64--70.

\bibitem{softPneu}
F.~Ni, D.~Rojas, K.~Tang, L.~Cai, and T.~Asfour, ``A jumping robot using soft pneumatic actuator,'' in \emph{2015 IEEE International Conference on Robotics and Automation (ICRA)}, 2015, pp. 3154--3159.

\bibitem{marc3Dhop}
M.~H. Raibert, J.~H.~Benjamin~Brown, and M.~Chepponis, ``Experiments in balance with a 3d one-legged hopping machine,'' \emph{The International Journal of Robotics Research}, vol.~3, no.~2, pp. 75--92, 1984.

\bibitem{festoBionicKang}
K.~Graichen, S.~Hentzelt, A.~Hildebrandt, N.~Kärcher, N.~Gaißert, and E.~Knubben, ``Control design for a bionic kangaroo,'' \emph{Control Engineering Practice}, vol.~42, pp. 106--117, 2015.

\bibitem{hydExo}
B.~Song, D.~Lee, S.~Y. Park, and Y.~S. Baek, ``Design and performance of nonlinear control for an electro-hydraulic actuator considering a wearable robot,'' \emph{Processes}, vol.~7, no.~6, 2019.

\bibitem{surveyHy}
J.~Mattila, J.~Koivumäki, D.~G. Caldwell, and C.~Semini, ``A survey on control of hydraulic robotic manipulators with projection to future trends,'' \emph{IEEE/ASME Transactions on Mechatronics}, vol.~22, no.~2, pp. 669--680, 2017.

\bibitem{salto}
D.~W. Haldane, J.~K. Yim, and R.~S. Fearing, ``Repetitive extreme-acceleration (14-g) spatial jumping with salto-1p,'' in \emph{2017 IEEE/RSJ International Conference on Intelligent Robots and Systems (IROS)}, 2017, pp. 3345--3351.

\bibitem{TAUB}
V.~Zaitsev, O.~Gvirsman, U.~Ben~Hanan, A.~Weiss, A.~Ayali, and G.~Kosa, ``Locust-inspired miniature jumping robot,'' in \emph{2015 IEEE/RSJ International Conference on Intelligent Robots and Systems (IROS)}, 2015, pp. 553--558.

\bibitem{jumpROACH}
G.-P. Jung, C.~S. Casarez, J.~Lee, S.-M. Baek, S.-J. Yim, S.-H. Chae, R.~S. Fearing, and K.-J. Cho, ``Jumproach: A trajectory-adjustable integrated jumping–crawling robot,'' \emph{IEEE/ASME Transactions on Mechatronics}, vol.~24, no.~3, pp. 947--958, 2019.

\bibitem{grillo3}
F.~Li, W.~Liu, X.~Fu, G.~Bonsignori, U.~Scarfogliero, C.~Stefanini, and P.~Dario, ``Jumping like an insect: Design and dynamic optimization of a jumping mini robot based on bio-mimetic inspiration,'' \emph{Mechatronics}, vol.~22, no.~2, pp. 167--176, 2012.

\bibitem{ambrose2022creating}
E.~R. Ambrose, \emph{Creating ARCHER: A 3D Hopping Robot with Flywheels for Attitude Control}.\hskip 1em plus 0.5em minus 0.4em\relax California Institute of Technology, 2022.

\bibitem{suzuki2022ramiel}
T.~Suzuki, Y.~Toshimitsu, Y.~Nagamatsu, K.~Kawaharazuka, A.~Miki, Y.~Ribayashi, M.~Bando, K.~Kojima, Y.~Kakiuchi, K.~Okada, \emph{et~al.}, ``Ramiel: A parallel-wire driven monopedal robot for high and continuous jumping,'' in \emph{2022 IEEE/RSJ International Conference on Intelligent Robots and Systems (IROS)}.\hskip 1em plus 0.5em minus 0.4em\relax IEEE, 2022, pp. 5017--5024.

\bibitem{grimes2012design}
J.~A. Grimes and J.~W. Hurst, ``The design of atrias 1.0 a unique monopod, hopping robot,'' in \emph{Adaptive Mobile Robotics}.\hskip 1em plus 0.5em minus 0.4em\relax World Scientific, 2012, pp. 548--554.

\bibitem{cheeta3jump}
Q.~Nguyen, M.~J. Powell, B.~Katz, J.~D. Carlo, and S.~Kim, ``Optimized jumping on the mit cheetah 3 robot,'' in \emph{2019 International Conference on Robotics and Automation (ICRA)}, 2019, pp. 7448--7454.

\bibitem{spaceBOKjump}
H.~Kolvenbach, E.~Hampp, P.~Barton, R.~Zenkl, and M.~Hutter, ``Towards jumping locomotion for quadruped robots on the moon,'' in \emph{2019 IEEE/RSJ International Conference on Intelligent Robots and Systems (IROS)}, 2019, pp. 5459--5466.

\bibitem{minitaur}
G.~Kenneally, A.~De, and D.~E. Koditschek, ``Design principles for a family of direct-drive legged robots,'' \emph{IEEE Robotics and Automation Letters}, vol.~1, no.~2, pp. 900--907, 2016.

\bibitem{KURMET}
Y.~Liu, P.~M. Wensing, D.~E. Orin, and J.~P. Schmiedeler, ``Fuzzy controlled hopping in a biped robot,'' in \emph{2011 IEEE International Conference on Robotics and Automation}, 2011, pp. 1982--1989.

\bibitem{hoppy}
J.~Ramos, Y.~Ding, Y.-W. Sim, K.~Murphy, and D.~Block, ``Hoppy: An open-source kit for education with dynamic legged robots,'' in \emph{2021 IEEE/RSJ International Conference on Intelligent Robots and Systems (IROS)}, 2021, pp. 4312--4318.

\bibitem{BRL}
J.~Babič, B.~Lim, D.~Omrčen, J.~Lenarčič, and F.~C. Park, ``{A Biarticulated Robotic Leg for Jumping Movements: Theory and Experiments},'' \emph{Journal of Mechanisms and Robotics}, vol.~1, no.~1, p. 011013, 08 2008.

\bibitem{Ascento}
V.~Klemm, A.~Morra, C.~Salzmann, F.~Tschopp, K.~Bodie, L.~Gulich, N.~Küng, D.~Mannhart, C.~Pfister, M.~Vierneisel, F.~Weber, R.~Deuber, and R.~Siegwart, ``Ascento: A two-wheeled jumping robot,'' in \emph{2019 International Conference on Robotics and Automation (ICRA)}, 2019, pp. 7515--7521.

\bibitem{cassieRL}
Z.~Li, X.~B. Peng, P.~Abbeel, S.~Levine, G.~Berseth, and K.~Sreenath, ``Robust and versatile bipedal jumping control through multi-task reinforcement learning,'' \emph{arXiv preprint arXiv:2302.09450}, 2023.

\bibitem{Unitree}
``Unitree h1 jumping video,'' \url{https://www.youtube.com/watch?v=83ShvgtyFAg}.

\bibitem{hyperleg}
\BIBentryALTinterwordspacing
Y.-J. Kim, D.-Y. Kim, C.-Y. Maeng, and S.-H. Yun, ``Introducing hyperleg: Human-like robot leg and foot for highly dynamic motions,'' May 2023. [Online]. Available: \url{https://www.youtube.com/watch?v=wLFCMwRvhVI}
\BIBentrySTDinterwordspacing

\bibitem{tello}
Y.~Sim and J.~Ramos, ``Tello leg: The study of design principles and metrics for dynamic humanoid robots,'' \emph{IEEE Robotics and Automation Letters}, vol.~7, no.~4, pp. 9318--9325, 2022.

\bibitem{chou2009role}
S.-W. Chou, H.-Y.~K. Cheng, J.-H. Chen, Y.-Y. Ju, Y.-C. Lin, and M.-K.~A. Wong, ``The role of the great toe in balance performance,'' \emph{Journal of Orthopaedic Research}, vol.~27, no.~4, pp. 549--554, 2009.

\bibitem{toeathletic}
J.-P. Goldmann, M.~Sanno, S.~Willwacher, K.~Heinrich, and G.-P. Brüggemann, ``The potential of toe flexor muscles to enhance performance,'' \emph{Journal of Sports Sciences}, vol.~31, no.~4, pp. 424--433, 2013.

\bibitem{wang2006influence}
L.~Wang, Z.~Yu, Q.~Meng, and Z.~Zhang, ``Influence analysis of toe-joint on biped gaits,'' in \emph{2006 International Conference on Mechatronics and Automation}.\hskip 1em plus 0.5em minus 0.4em\relax IEEE, 2006, pp. 1631--1635.

\bibitem{ogura2006human}
Y.~Ogura, K.~Shimomura, H.~Kondo, A.~Morishima, T.~Okubo, S.~Momoki, H.-o. Lim, and A.~Takanishi, ``Human-like walking with knee stretched, heel-contact and toe-off motion by a humanoid robot,'' in \emph{2006 IEEE/RSJ International Conference on Intelligent Robots and Systems}.\hskip 1em plus 0.5em minus 0.4em\relax IEEE, 2006, pp. 3976--3981.

\bibitem{hashimoto2010study}
K.~Hashimoto, Y.~Takezaki, K.~Hattori, H.~Kondo, T.~Takashima, H.-o. Lim, and A.~Takanishi, ``A study of function of foot's medial longitudinal arch using biped humanoid robot,'' in \emph{2010 IEEE/RSJ International Conference on Intelligent Robots and Systems}.\hskip 1em plus 0.5em minus 0.4em\relax IEEE, 2010, pp. 2206--2211.

\bibitem{sato2010trajectory}
T.~Sato, S.~Sakaino, and K.~Ohnishi, ``Trajectory planning and control for biped robot with toe and heel joints,'' in \emph{2010 11th IEEE International Workshop on Advanced Motion Control (AMC)}.\hskip 1em plus 0.5em minus 0.4em\relax IEEE, 2010, pp. 129--136.

\bibitem{choi2016design}
W.~Choi, G.~A. Medrano-Cerda, D.~G. Caldwell, and N.~G. Tsagarakis, ``Design of a variable compliant humanoid foot with a new toe mechanism,'' in \emph{2016 IEEE International Conference on Robotics and Automation (ICRA)}.\hskip 1em plus 0.5em minus 0.4em\relax IEEE, 2016, pp. 642--647.

\bibitem{yamamoto2007toe}
K.~Yamamoto, T.~Sugihara, and Y.~Nakamura, ``Toe joint mechanism using parallel four-bar linkage enabling humanlike multiple support at toe pad and toe tip,'' in \emph{2007 7th IEEE-RAS International Conference on Humanoid Robots}.\hskip 1em plus 0.5em minus 0.4em\relax IEEE, 2007, pp. 410--415.

\bibitem{NASAanthro}
``Anthropometry and biomechanics,'' \url{https://msis.jsc.nasa.gov/sections/section03.htm}.

\bibitem{ANSYS}
``Ansys topology optimization,'' \url{https://www.ansys.com/applications/topology-optimization}.

\bibitem{passiveankle}
D.~Kim, S.~J. Jorgensen, J.~Lee, J.~Ahn, J.~Luo, and L.~Sentis, ``Dynamic locomotion for passive-ankle biped robots and humanoids using whole-body locomotion control,'' \emph{The International Journal of Robotics Research}, vol.~39, no.~8, pp. 936--956, 2020.

\bibitem{chignoli2021humanoid}
M.~Chignoli, D.~Kim, E.~Stanger-Jones, and S.~Kim, ``The mit humanoid robot: Design, motion planning, and control for acrobatic behaviors,'' in \emph{2020 IEEE-RAS 20th International Conference on Humanoid Robots (Humanoids)}.\hskip 1em plus 0.5em minus 0.4em\relax IEEE, 2021, pp. 1--8.

\bibitem{dai}
H.~Dai, A.~Valenzuela, and R.~Tedrake, ``Whole-body motion planning with centroidal dynamics and full kinematics,'' in \emph{2014 IEEE-RAS International Conference on Humanoid Robots}, 2014, pp. 295--302.

\bibitem{kim2019highly}
D.~Kim, J.~D. Carlo, B.~Katz, G.~Bledt, and S.~Kim, ``Highly dynamic quadruped locomotion via whole-body impulse control and model predictive control,'' \emph{arXiv: 1909.06586}, 2019.

\end{thebibliography}

\end{document}